\documentclass[letterpaper, 10 pt, conference]{ieeeconf}  

\IEEEoverridecommandlockouts                              

\usepackage{epsfig} 
\usepackage{mathptmx} 
\usepackage{times} 
\usepackage{amsmath} 

\usepackage{graphics} 
\usepackage{graphicx} 
\usepackage{underscore} 
\usepackage{color} 
\usepackage{booktabs} 

\usepackage{bbding}
\usepackage{pifont}
\usepackage{wasysym}
\usepackage[normalem]{ulem} 
\usepackage{xcolor} 
\usepackage{hyperref}

\newcommand{\blah}{blah blah blah blah blah blah blah blah blah blah blah blah blah blah blah blah blah blah blah blah blah blah blah blah blah blah blah blah blah blah}

\title{\LARGE \bf
Map-based Modular Approach for Zero-shot Embodied Question Answering
}

\author{Koya Sakamoto$^{1}$, Daichi Azuma$^{2}$, Taiki Miyanishi$^{3,4,6}$, Shuhei Kurita$^{5,6}$ and Motoaki Kawanabe$^{4,6}$
\thanks{$^{1}$ Graduate School of Informatics, Kyoto University, Japan}%
\thanks{$^{2}$ Sony Semiconductor Solutions}%
\thanks{$^{3}$ The University of Tokyo, Japan}%
\thanks{$^{4}$ Advanced Telecommunications Research Institute International (ATR), Kyoto, Japan}%
\thanks{$^{5}$ National Institute of Informatics, Tokyo, Japan}%
\thanks{$^{6}$ RIKEN Center for Advanced Intelligence Project, Tokyo, Japan}%
}

\begin{document}

\maketitle
\thispagestyle{empty}
\pagestyle{empty}

\begin{abstract}

Embodied Question Answering (EQA) serves as a benchmark task to evaluate the capability of robots to navigate within novel environments and identify objects in response to human queries. 
However, existing EQA methods often rely on simulated environments and operate with limited vocabularies. 
This paper presents a map-based modular approach to EQA, enabling real-world robots to explore and map unknown environments. 
By leveraging foundation models, our method facilitates answering a diverse range of questions using natural language. 
We conducted extensive experiments in both virtual and real-world settings, demonstrating the robustness of our approach in navigating and comprehending queries within unknown environments.
(\href{https://atr-dbi.github.io/Map-EQA/}{Webpage})
\end{abstract}

\section{INTRODUCTION}
Home robots that interact with humans need to understand both language and 3D environments to perform household tasks based on human instructions. 
For instance, if we misplace our smartphones, it would be helpful if robots could search the room and locate them for us. 
To accomplish this, robots must explore scenes to find the target object and generate text-based responses based on their visual observations. 
These skills can be assessed through the Embodied Question Answering (EQA) task~\cite{embodiedqa, EqaMatterport}, where an agent navigates an unfamiliar environment to answer a question.
Recent studies of semantic visual navigation~\cite{navigatingtoobjects} indicate that modular learning approaches are effective in real-world scenarios, whereas end-to-end learning approaches fail due to a significant domain gap in visual observations between simulations and reality.
The existing EQA methods~\cite{embodiedqa, EqaMatterport} utilize an end-to-end framework trained on simulation environments, which is likely to lead to poor real-world performance.
In addition, the VQA modules of the existing methods often struggle to deal with new types of questions and new objects because the models are trained with a limited vocabulary and a few question types. 
In terms of question diversity, the MP3D-EQA dataset suffers from a limited range of question types, primarily focusing on ``what color'' or ``what room'' even though in real-world situations, we ask a wider variety of questions including ``where'' and ``what is''.
When we consider EQA in more realistic daily life settings, EQA models need to be able to handle an open vocabulary and a diverse range of questions.

\begin{figure}[t]
  \centering
  \includegraphics[keepaspectratio, scale=0.26]{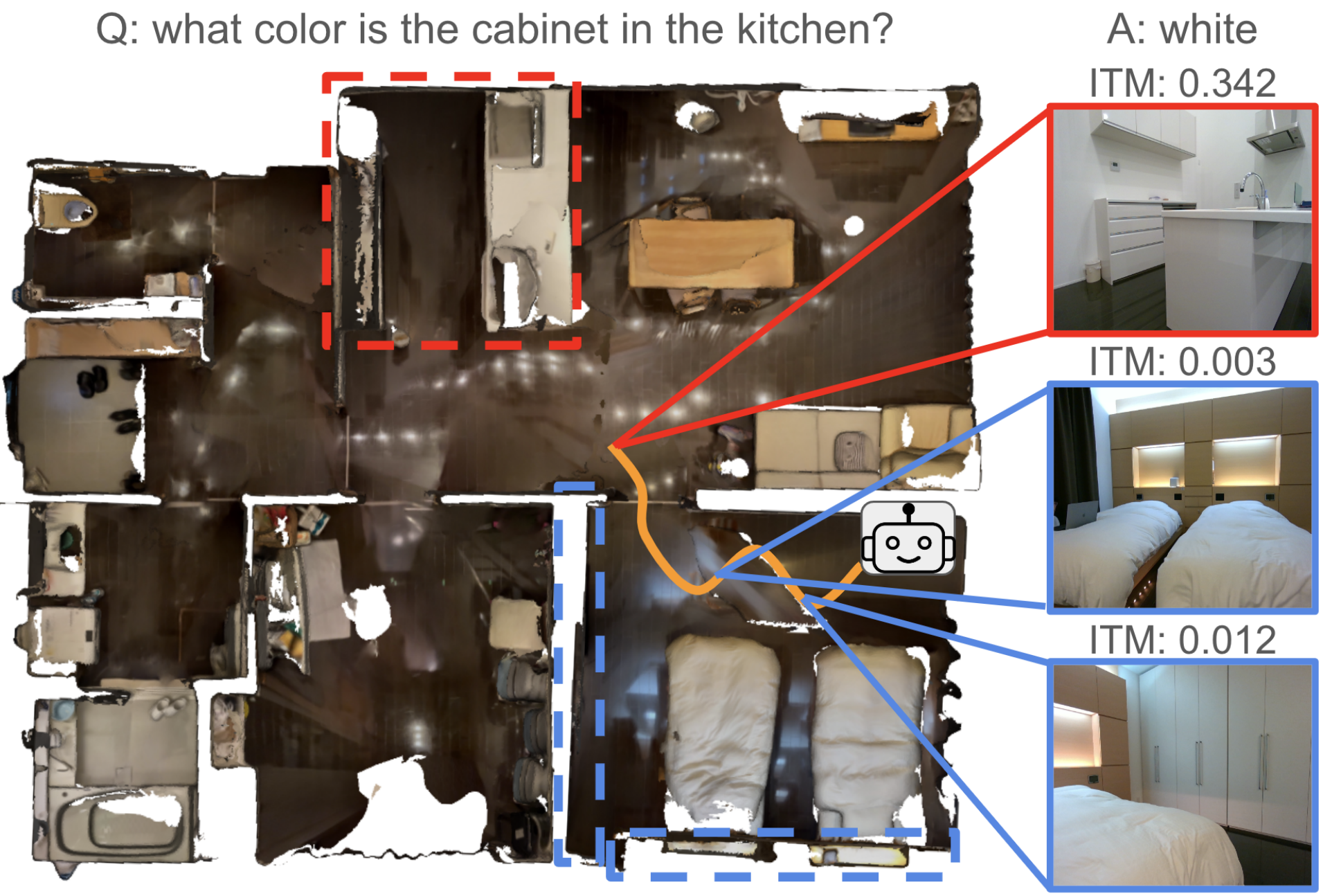}
  \caption{Example of our method: We provide an agent with a question and the agent proceeds to explore an unknown environment. When it encounters a potential target object, it verifies if this is indeed the correct target object through image-text matching (ITM). If the ITM score falls below a pre-determined threshold, the agent continues its exploration. If the ITM score exceeds the threshold, the agent stops exploration and performs VQA.}
  \label{figurelabel}
  \vspace{-0.3cm}
\end{figure}

In this paper, we present a map-based modular EQA method combining object-goal navigation (ObjNav)~\cite{chaplot2020object, ObjnavImplicitMaps, ramakrishnan2022poni} and Visual Question Answering (VQA)~\cite{VQA, li2022blip, li2023blip2, liu2023llava} tasks.
The agent extracts a target object category from a question, explores the environment using frontier-based exploration~\cite{FrontierExploration}, and answers the question with an open vocabulary once the target object is found. 
Unlike end-to-end EQA methods relying on reinforcement learning, our approach is designed to work robustly in real-world scenarios.
\if[]
To address the scarcity of realistic EQA datasets, we also proposed a ScanNet-EQA dataset based on the popular 3D indoor dataset of ScanNet~\cite{dai2017scannet}. 
Unlike the previous EQA dataset based on Matterport 3D (MP3D), our ScanNet-EQA dataset contains 36,366 human-annotated question-and-answer pairs about the 675 scenes from the ScanQA dataset~\cite{Azuma2022CVPR}.
\fi

\if[]
Our motivation is to address accidental obstacle collision problems and closed vocabulary sets, which may not be simply solved by previous reinforcement learning methods. In order to solve these problems, we utilize other methods on corresponding tasks such as Visual Question Answering (VQA)~\cite{VQA} and object goal navigation. The existing and public dataset~\cite{EqaMatterport} cannot measure our method's applicability as its vocabularies of both questions and answers are limited. Thus, we created a new dataset on ScanNet~\cite{dai2017scannet}. The vocabulary sizes of questions and answers are over 1000. 
\fi

\if[]
We introduce our method based on an assumption that EQA is composed of three auxiliary tasks: target object extraction, object goal navigation, and visual question answering. We humans also search for objects to answer questions in unseen environments. In our framework, an agent is given a question and extract a target object category from the question. The agent searches for objects of the extracted object category. When it finds a target object, it determines whether the object is something to do with the question. If it is related to question, the agent does VQA.
\fi

%
%

Using the MP3D-EQA dataset, we evaluate our proposed method and find that it performs comparably or even outperforms existing end-to-end methods that employ reinforcement learning. 
On MP3D-EQA, the VQA top-1 accuracy scores around 0.43, which is higher than the scores stated on existing methods~\cite{EqaMatterport}. 
%
We also conduct extensive surveys in two real houses, using question formats that differ from those in MP3D-EQA dataset and incorporating target objects not present in MP3D-EQA for our experiments.
The results demonstrated that our map-based modular approach achieved high question-answer accuracy.




\section{RELATED WORK}

\subsection{Visual Question Answering in 3D Space}
In the domain of 3D spatial understanding and question answering (3D-QA)~\cite{Azuma2022CVPR,3dllm,ye20223d}, models provide answers to textual questions regarding rich RGB-D indoor scans encapsulating entire 3D scenes. 
Distinguished from conventional 2D-QA~\cite{VQA} models commonly employed in visual question answering, the 3D-QA task poses distinct challenges, particularly concerning spatial comprehension, object alignment, directionality, and localization based on textual cues within a 3D setting. 
However, environments change dynamically, and it is necessary to gather information anew each time.
Thus, EQA is a more realistic task as an agent explores an unseen environment and answers a question.

\subsection{Language-Guided Object Goal Navigation}
Object goal navigation is a task in which an embodied agent follows a concise textual phrase specifying a target object category, navigates through a 3D environment, and finally locates and reach the target object~\cite{chaplot2020object}.
It also shares similarities with the vision and language navigation (VLN) task, in which agents follow detailed navigation instructions~\cite{vlnAnderson2018CVPR,vlnce,vlnpano2real}.
using the noisy-channel language modelings~\cite{glgp} or a history-aware multimodal transformer~\cite{hamt}.
In object goal navigation, SemExp~\cite{chaplot2020object} and PONI~\cite{ramakrishnan2022poni} adopt map-based approaches. They generate semantic maps by utilizing semantic segmentation and top-down projection. These maps are then leveraged to determine long-term goals and actions for the agent.
%
In both methods, they develop global policies to infer long-term goals with limited vocabularies. 
Consequently, these methods cannot handle many other object categories that are not included in the training data.
To address the vocabulary limitation, zero-shot object goal navigation methods have been proposed~\cite{majumdar2022zson, gadre2022cow, chang2023goat}, which enable navigation to objects even if they were not explicitly encountered during training.
GOAT~\cite{chang2023goat} demonstrates the ability to navigate to any object or location using text, images, or object categories.

\subsection{Embodied Referring Expression Comprehension} 
Referring expression comprehension is a task for localizing objects following a short textual phrase of the referring expression and introduced in 2D images~\cite{referitgame, flickr30k,refcoco,refcocog}.
Previous work has explored the use of referring expressions for embodied agents~\cite{Qi2020CVPR}. 
Additionally, research has been conducted on tasks requiring both referring expression comprehension and object manipulation for these agents~\cite{sima2022embodiedReferringExpressionManipulation}. 
Furthermore, referring expression comprehension has been applied to first-person video settings, which closely resembles the robotic navigation context~\cite{refego}.

\subsection{Question Answering for Embodied Agents} 
Embodied AI agents equipped with VQA can analyze visual input from cameras or other sensors to answer questions about their environment.
To assist the agent in recognizing and comprehending the scene during navigation, VQA is used~\cite{Kawaharazuka2023}.
For example, by answering a question about its surroundings, the agent can prevent collisions with transparent doors.
In EQA, a crucial challenge is VQA after navigation, where an agent explores an unseen environment to answer a given question.
The original EQA~\cite{embodiedqa} and MP3D-EQA~\cite{EqaMatterport} datasets contain questions that almost focus on a single target within the House3D~\cite{wu2018building} and Matterport3D~\cite{Matterport3D} environments. 
A generalized version of EQA has been proposed~\cite{yu2019mteqa}, in which each question within this expanded task encompasses multiple objects, necessitating the agent to navigate to these objects to provide an answer.
K-EQA~\cite{KnowledgebasedEqa} presents a dataset where questions (e.g. ``Please tell me what objects are used to cut food in the room?'') require prior knowledge such as ``knife is used for cutting food''.
In these EQA tasks, end-to-end imitation learning approaches on shortest paths are often used as the baseline models~\cite{embodiedqa, EqaMatterport}.
However, this approach frequently results in collisions with walls, which is undesirable when deploying these models in real-world scenarios.
Another standard model~\cite{EqaMatterport} uses point cloud data so that the agent does not collide with the objects.
Unfortunately, both models are based on supervised methods and suffer from limited vocabularies.

\section{Proposed Method}
\begin{figure*}[t]
  \centering
  \includegraphics[keepaspectratio, width=0.95\textwidth]{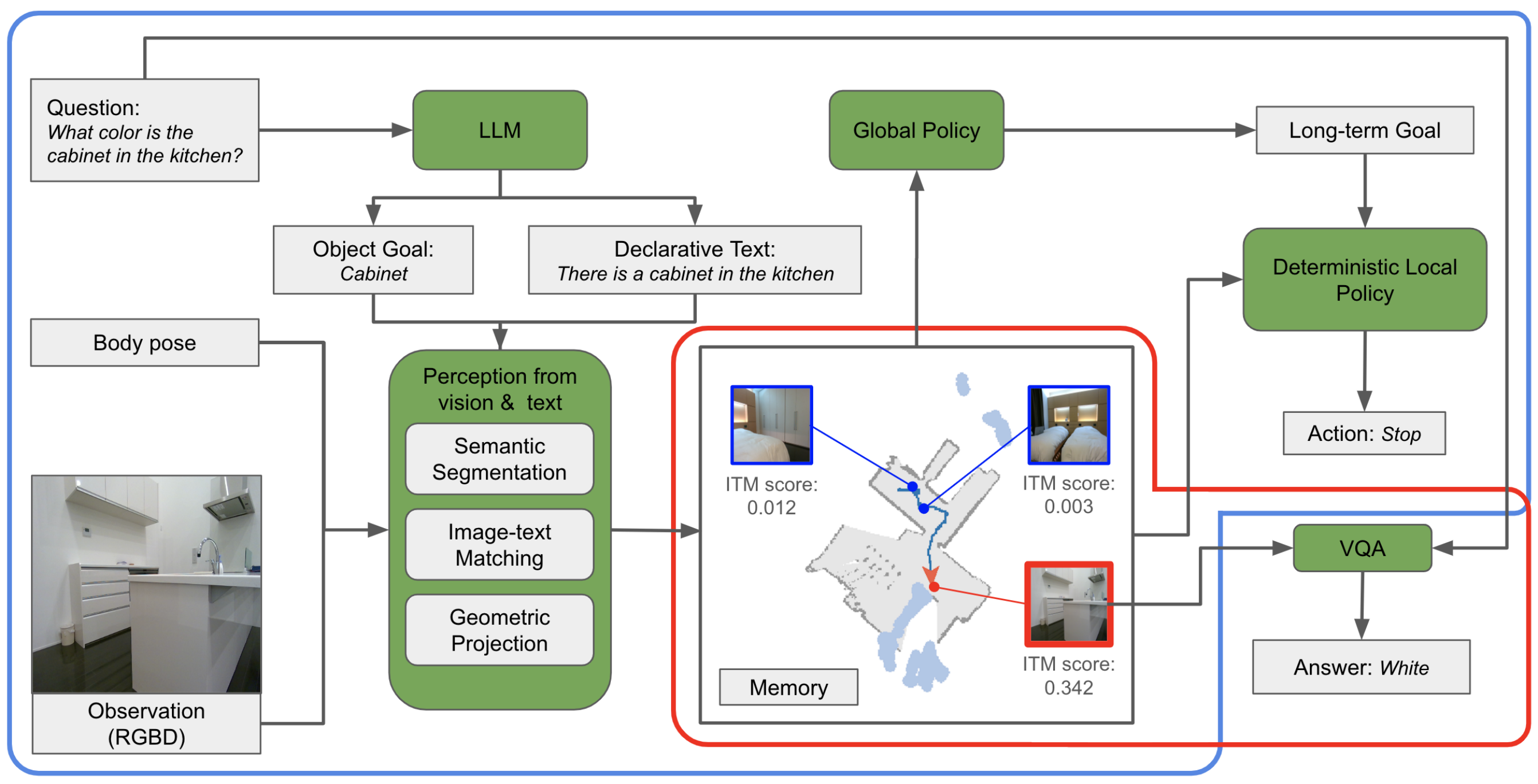}
  \caption{\textbf{
   Map-based Modular Embodied Question Answering Model Overview.} 
 The proposed method comprises the Navigation module (outlined in blue) and the VQA module (outlined in red).
 The Navigation module consists of the Perception module and a set of Policies. 
 The Perception module incrementally builds a 2D map, storing images along with their image-text matching scores. 
 The Global Policy selects a long-term goal based on the 2D map and its frontiers. 
 The Deterministic Local Policy outputs actions, and finally, the VQA module provides an answer based on the memorized images and the given question.
}
  \label{fig:OverviewProposedMethod}
\end{figure*}


\if[]
We describe the definition of the EQA task in Section~\ref{sec:Chap3TaskDef}.
In Section~\ref{sec:Chap3Overview}, we then show the overview of our modular approach followed by their modules in Section~\ref{}.
We introduce the ScanNet-EQA dataset comprising 36k QA pairs from ScanNet and their corresponding trajectories between start and end points in Section~\ref{}.
\fi


\subsection{Task Definition}
\label{sec:Chap3TaskDef}
The EQA task aims to answer a question by exploring and finding a target object in the unseen 3D world.
During EQA, an agent can observe its location, orientation, RGB-D image, and the given question from a user. 
The agent can take actions such as moving forward, turning left or right, and stopping.
After finding the target object, the agent performs VQA based on the observed images and the posed question.
The episode is considered successful if the predicted answer matches the correct answer.

\subsection{Overview of EQA Framework}
\label{sec:Chap3Overview}
As illustrated in Fig~\ref{fig:OverviewProposedMethod}, our EQA framework mainly consists of language-guided navigation and VQA modules.
First, an agent is placed within an unknown environment and is simultaneously given a question asking about the surroundings (e.g. ``What color are the cabinets in the kitchen?''). 
Next, the agent explores the indoor environment, observing the images from its egocentric view.
When the agent finds an object belonging to the extracted object category, it determines whether the object is the target object by image-text matching. If the object is considered the target object, the ``stop" action is selected and the navigation is terminated.
%
Finally, the VQA module predicts the answer based on the images collected up to the current state, taking into account the image-text matching scores.

\subsection{Language-guided Navigation Module}
\label{sec:Chap3NavModule}
\begin{figure}[t]
  \centering
  \includegraphics[keepaspectratio, width=0.45\textwidth]{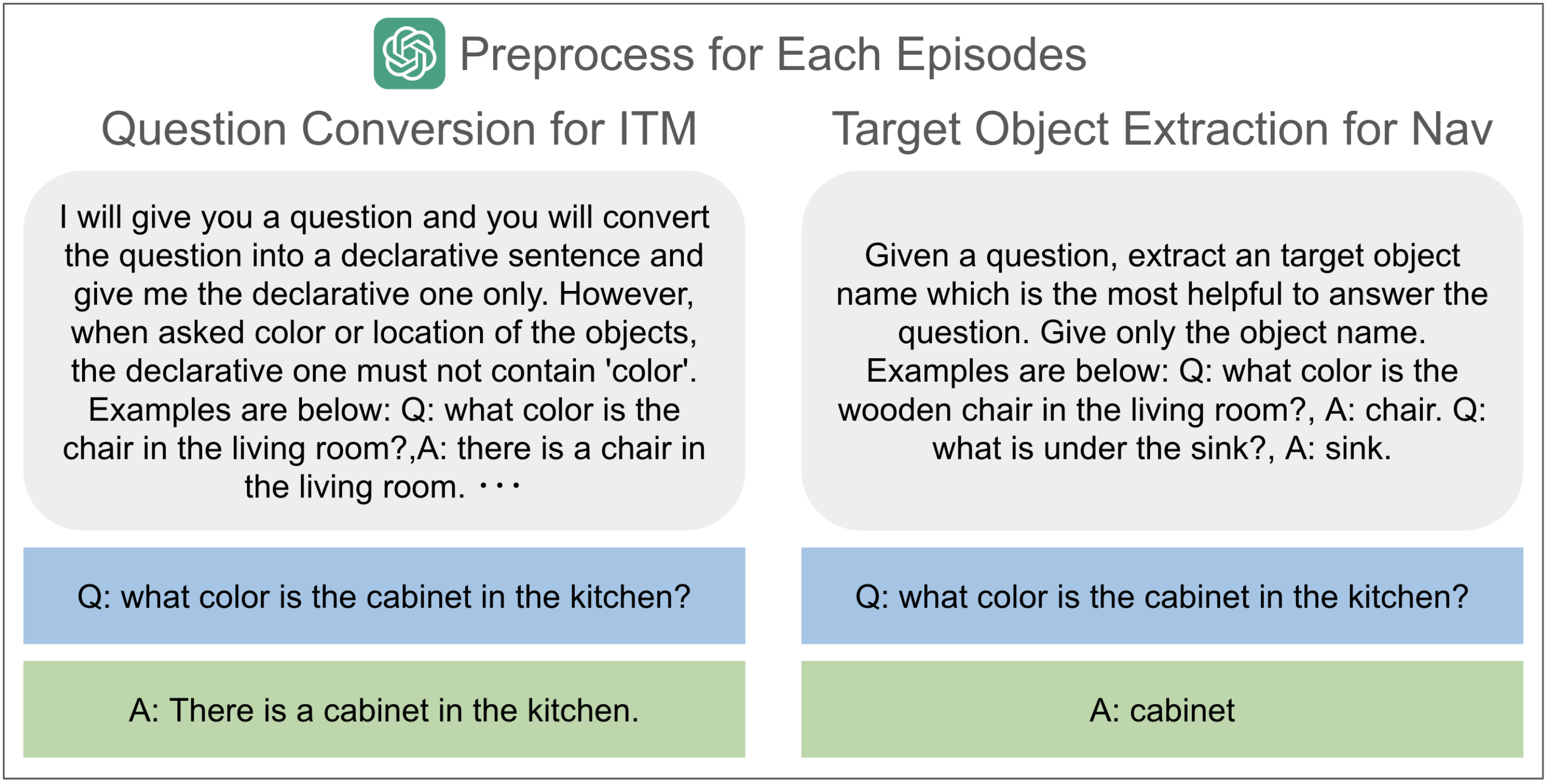}
  \caption{Dataset pre-processing using \texttt{gpt-35-turbo-0613}. It extracts a target object category from a given question for ObjNav and converts a question into a declarative text for image-text matching.}
  \label{fig:preprocess}
\end{figure}

\vspace{0.2cm}
\noindent \textbf{Language Understanding.}
Our navigation module detects the object category extracted from a given question and explores to find the corresponding target object.
We used the \texttt{gpt-35-turbo-0613} through the Azure OpenAI API to extract an object category with a prompt in Fig~\ref{fig:preprocess}.
We also use the \texttt{gpt-35-turbo-0613} for converting questions into declarative sentences for image-text matching (ITM).

\vspace{0.2cm}
\noindent \textbf{Scene Understanding.}
The scene understanding module generates a semantic map for navigation using object detection on first-person images.
We use Detic~\cite{detic}, which identifies 21,00 object classes, to segment observed images based on the object category extracted by Large Language Models.
Then, this module overlays the object detection outcomes with depth information and projects them onto a top-down view.


\if[]
\vspace{0.2cm}
\noindent \textbf{Semantic Fields}
\fi

\if[]
\vspace{0.2cm}
\noindent \textbf{Global Planner}
\blah

\vspace{0.2cm}
\noindent \textbf{Local Planner}
\fi

\vspace{0.2cm}
\noindent \textbf{Planner.}
To find the target object, we use frontier-based exploration~\cite{FrontierExploration}, which selects the closest unexplored region as a goal.
We assumed frontier-based exploration does not need training and, thus, results in a minimal domain gap between simulation and real-world performance.
Using the created semantic map, the agent first detects frontiers, defined as the edges or boundaries between known and unknown areas within an environment.
The closest frontier to the agent is designated as a long-term goal if multiple frontiers exist.
The agent uses the A* algorithm to determine the path between its current position and the long-term goal and selects a sequence of actions to move to that position.
The long-term goal is updated every 25 steps to simulate in parallel. 
As for the experiments in the real world, the updating steps change dynamically according to the observation of the target object and reaching the long-term goal.
After finding an object belonging to the extracted object category, the agent sets the target object's position to the long-term goal.
While exploring the environment, the agent stores images when it approaches the target object within one meter and faces the center position of the target object for the subsequent image-text matching and VQA modules.
Exploration stops after 100 steps or when a stop action is selected (the text-image matching score is greater than $\beta$, which will be introduced later.)

\subsection{Image-text Matching Module}
The agent has to distinguish target objects from others based on a declarative text converted from a question.
To tackle this problem, we use vision-language foundation models BLIP2~\cite{li2023blip2} and CLIP~\cite{clip} as an image-text matching module. 
Using these foundation models, we measure the similarity between the observed images and the declarative sentence.
We assumed that the similarity between the image containing the target object and the declarative sentence would be greater than the similarity between an image without the target object and the declarative sentence.
The agent stops and performs VQA on the image when the similarity score exceeds $\beta$, otherwise, it continues to move.

\subsection{Visual Question Answering Module}
After navigation, we obtain a set of images and select the one with the highest similarity score for VQA.
We use the pre-trained vision-language models for this VQA module such as BLIP~\cite{li2022blip}, BLIP2~\cite{li2023blip2}, and LLaVA~\cite{liu2023llava} respectively, which show high performance on many VQA tasks.


\section{EXPERIMENTS}

\subsection{EQA Datasets}
\vspace{0.2cm}
Our experiments leverage the Matterport3D (MP3D) EQA dataset within the Habitat simulator~\cite{habitat19iccv, szot2021habitat, puig2023habitat3}. 
The dataset uses scenes derived from 3D reconstructions of real-world settings. 
Since MP3D-EQA~\cite{EqaMatterport} only releases train and validation splits, we further divided the original training dataset into new train and val sets based on the scenes. 
The original validation dataset was then used as the test dataset.

The agent is equipped with sensors including an RGB-D sensor and a pose sensor. 
The observation space encompasses RGB-D images with dimensions $480*640$. 
The pose sensor reports the agent's position and rotation.
The agent is spawned at distances corresponding to 10, 30, and 50 actions away from the ground truth end positions, moving towards the start positions along the shortest paths.
The shortest path lengths from these start positions to end positions are 3.45, 4.53, 5.71, and 8.21 meters.

\subsection{Implementation Details}
We mainly use the implementation of SemExp~\cite{chaplot2020object}.
There are two hyperparameters: the thresholds $\alpha, \beta$ of object detection and image-text matching.
We set $(\alpha, \beta) \in {(0.3, 0.0), (0.2, 0.1), (0.1, 0.2)}$.
\if[]
\textcolor{red}{\uline{
Lower $\alpha$ means that an object detection model is likely to detect a target object and other objects as well.
Higher $\beta$ means that the agent performs VQA on the image with the highest image-text matching score.
}}
\fi
A lower value for $\alpha$ implies that the object detection model is more likely to detect not only the target object but also other objects present in the scene.
A higher value for $\beta$ indicates that the agent prioritizes performing VQA on the image with the highest image-text matching score, suggesting greater confidence in that particular image's relevance to the question.


\begin{figure}[h]
  \centering
  \includegraphics[keepaspectratio, width=0.4\textwidth]{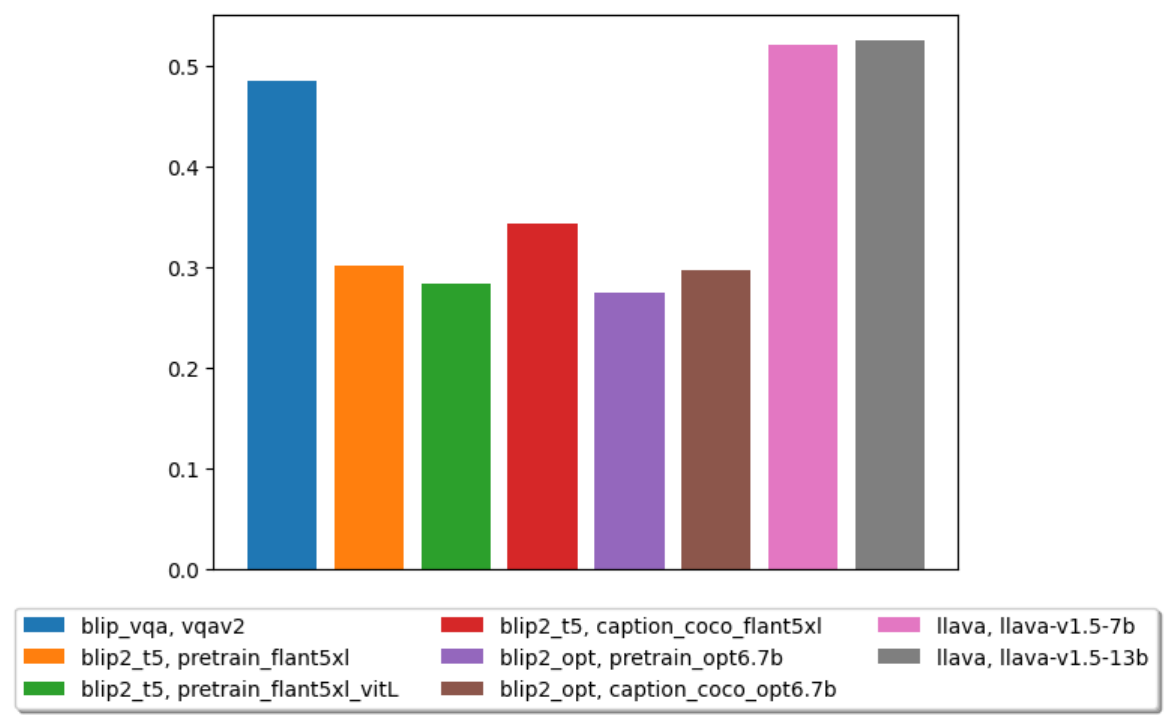}
  \caption{VQA top-1 accuracy on MP3D-EQA train'. The scores of \texttt{LLaVA-v1.5-7b} and \texttt{LLaVA-v1.5-13b} are higher than those of others.}
  \label{fig:ResultVQA}
\end{figure}




\subsection{Evaluation Metrics}
We use the following metrics for evaluation: VQA top-1 accuracy, \textbf{$d_{T}$} (distance to target), 
following previous works~\cite{EqaMatterport,embodiedqa}.
The VQA top-1 accuracy is defined as the rate at the output of the VQA model with the highest probability matches the ground truth answer.
The $\mathbf{d_{T}}$ is defined as the distance to the target from the agent position along the shortest path.
In our setting, the target position is defined as the agent end position of the shortest paths. 
As for our methods, the $\mathbf{d_{T}}$ is calculated between the target position and the position on which image-text matching scores higher than $\beta$ or the highest. 
These metrics are calculated for {$\mathbf{T_{-N}}$}, which is defined as a start position. 
According to ground truth shortest paths, we set an agent on $N$ back steps away from an end position. 
We can investigate how well our navigation module works by comparing $\mathbf{d_{T}}$ with distance to a target from a start position or comparing VQA top-1 accuracy of our method with that of VQA only experiments.

\vspace{0.2cm}
\subsection{Image-Text Matching}
The agent has to identify which object is a target object.
To enhance EQA accuracy and prevent the misidentification of irrelevant objects, we conducted extensive experiments to identify the most effective combination of models and caption formats.


\subsection{Experimental Results}
\begin{figure}[t]
  \centering
  \includegraphics[keepaspectratio, width=0.43\textwidth]{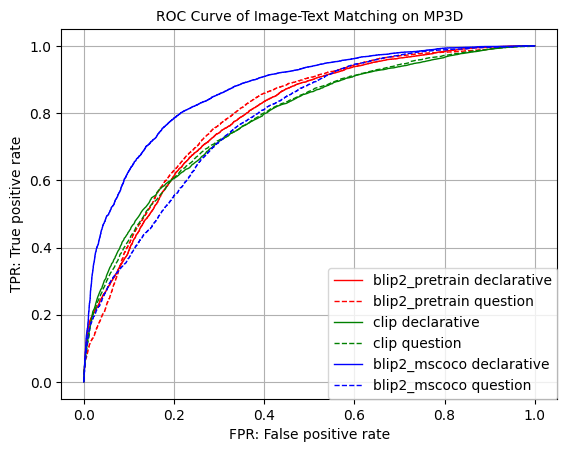}
  \caption{ROC Curves of Image-text Matching of MP3D-EQA \cite{EqaMatterport} at `train' split.
  }
  \label{fig:ITMScores}
\end{figure}

\vspace{0.2cm}
\noindent 
\textbf{Image-Text Matching.}
In Figure~\ref{fig:ITMScores}, we compare scores of \texttt{BLIP2-pretrain}, \texttt{BLIP2-MSCOCO}, and \texttt{CLIP}. \texttt{BLIP2-MSCOCO} is a BLIP2 model fine-tuned on MSCOCO~\cite{mscoco}.
The combination of declarative text and \texttt{BLIP2-MSCOCO} scores higher than the others on the MP3D-EQA dataset.
This suggests that the MSCOCO dataset [mscoco] might share similarities with the dataset used for this EQA task. 
Therefore, we will employ \texttt{BLIP2-MSCOCO} for the image-text matching module in our simulation experiments.

\vspace{0.2cm}
\noindent 
\textbf{VQA Baseline.}
We compare our method with the VQA method where the agent perform VQA at initial starting positions.
Figure~\ref{fig:ResultVQA} illustrates that \texttt{LLaVA-v1.5-7b} and \texttt{13b} score higher than others. 
It is known that LLaVA generally outperforms BLIP2 in various vision-language ~\cite{fu2023mme} and the results presented in Fig.~\ref{fig:ResultVQA} appear to align with this observation.
Considering the results, we adopted \texttt{LLaVA-v1.5-7b} for the experiment in the real-world environment.


\vspace{0.2cm}
\noindent \textbf{EQA in the Simulation Environment.}
\if[]
\begin{table*}[t]
    \centering
    \caption{Navigation and VQA scores on Scannet EQA. }
    \begin{tabular}{|c|cccc|cccc|}
        \hline
        \multicolumn{1}{|c|}{Method} &
        \multicolumn{4}{c|}{Navigation} &
        \multicolumn{4}{c|}{QA}\\
        \empty & 
        \multicolumn{4}{c|}{$d_{T}$ $\downarrow$} & \multicolumn{4}{c|}{Top-1 $\uparrow$}\\
        \empty & $T_{-10}$ & $T_{-30}$ & $T_{-50}$ & random & $T_{-10}$ & $T_{-30}$ & $T_{-50}$ & random\\
        \hline
        VQA only &  &  &  &  & 
        0.1334 & 0.1392 & 0.1376 & 0.138 \\
        $\alpha=0.3, \beta=0.0$ & 1.472 & 1.608 & 1.74 & 1.883 & 
        0.1463 & \textbf{0.1564} & \textbf{0.1539} & \textbf{0.1524} \\
        $\alpha=0.2, \beta=0.1$ & 1.501 & 1.617 & 1.758 & 1.878 & 
        0.147 & 0.154 & 0.1506 & 0.1514 \\
        $\alpha=0.1, \beta=0.2$ & \textbf{1.264} & \textbf{0.1555} & \textbf{1.618} & \textbf{1.773} & 
        \textbf{0.1501} & 0.1512 & 0.1492 & 0.1472 \\
        \hline
    \end{tabular}
    \label{tab:ScannetEqaResult}
\end{table*}
\fi
\begin{table*}[t]
    \centering
    \caption{Results on MP3D-EQA~\cite{EqaMatterport}. 
    We use \texttt{BLIP2-MSCOCO} as an image-text matching model and \texttt{LLAVA-v1.5-7b} as a VQA model. 
    }
    \begin{tabular}{|c|cccc|cccc|}
        \hline
        \multicolumn{1}{|c|}{Method} &
        \multicolumn{4}{c|}{Navigation} &
        \multicolumn{4}{c|}{QA}\\
        \empty & 
        \multicolumn{4}{c|}{$d_{T}$ $\downarrow$} & \multicolumn{4}{c|}{Top-1 $\uparrow$}\\
        \empty & $T_{-10}$ & $T_{-30}$ & $T_{-50}$ & random & $T_{-10}$ & $T_{-30}$ & $T_{-50}$ & random\\
        \hline
        VQA only (w/o navigation) & 3.45 & 4.53 & 5.71 & 8.21 & 0.383 & 0.389 & 0.327 & 0.305 \\
        Ours ($\alpha=0.3, \beta=0.0$) & 3.62 & 4.13 & 4.74 & 7.89 
        & 0.434 & 0.417 & 0.403 & 0.347 \\
        Ours ($\alpha=0.2, \beta=0.1$) & 3.60 & 4.13 & 4.70 & 7.80 
        & \textbf{0.450} & 0.429 & \textbf{0.418} & 0.358 \\
        Ours ($\alpha=0.1, \beta=0.2$) & \textbf{3.39} & \textbf{3.94} & \textbf{4.66} & \textbf{7.73} 
        & 0.445 & \textbf{0.434} & 0.409 & \textbf{0.368} \\
        \hline
    \end{tabular}
    \label{tab:MP3DEqaResult}
\end{table*}

We first measure the execution time for the EQA task.
The average time required to complete one episode, excluding any data pre-processing are 40.74, 50.39, and 62.07 seconds with $(\alpha, \beta)=(0.3,0.0), (0.2,0.1), (0.1,0.2)$ at random start positions respectively.
We then conduct the quantitative experiments reported in Table~\ref{tab:MP3DEqaResult}.
The results highlight our method consistently outperforms the VQA-only baseline.
It indicates that the navigation module of our method can work efficiently to answer questions.
We observe that the distance to the target $d_{T}$ using our method with $(\alpha, \beta)=(0.1,0.2)$ is shorter compared to the distances in cases where navigation is not employed.
However, the observed distance is significantly greater than the predefined stop distance of 1 meter. 
One potential reason for this discrepancy is that the agent might be navigating in the wrong direction, and cannot find the target objects.
We also observe that the object detection is segmenting part of an object located in front of the actual target. 
This misidentification might lead the agent to erroneously stop before reaching the intended destination.
The navigation scores and VQA top-1 accuracy of $(\alpha, \beta)=(0.1,0.2)$ predominantly higher than other combinations.
This outcome suggests that using a lower value for $\alpha$ and a higher value for $\beta$  leads to better performance.
In this scenario, VQA tends to be performed on the image with the highest image-text matching score.

\begin{table}[h]
    \centering
    \caption{Success and collisions in the real houses. We use \texttt{BLIP2-pretrain} and \texttt{BLIP2-MSCOCO} as an image-text matching model and \texttt{BLIP-pretrain}, \texttt{LLaVA-v1.5-7b} as a VQA model.}
    \begin{tabular}{|c|c|c|c|}
        \hline
        \multicolumn{1}{|c|}{ITM model} &
        \multicolumn{1}{c|}{VQA model} &
        \multicolumn{1}{c|}{success} &
        \multicolumn{1}{c|}{collisions}\\
        \hline
        \texttt{BLIP2-pretrain} & \texttt{BLIP-pretrain} & 7/19 & 3/19 \\
        \texttt{BLIP2-pretrain} & \texttt{LLaVA-v1.5-7b} & 6/19 & 3/19 \\
        \texttt{BLIP2-MSCOCO} & \texttt{BLIP-pretrain} & 6/19 & 4/19 \\
        \texttt{BLIP2-MSCOCO} & \texttt{LLaVA-v1.5-7b} & 6/19 & 4/19 \\
        \hline
    \end{tabular}
    \label{tab:RealWorld}
\end{table}

\begin{figure}[h]
  \centering
  \includegraphics[keepaspectratio, width=0.18\textwidth]{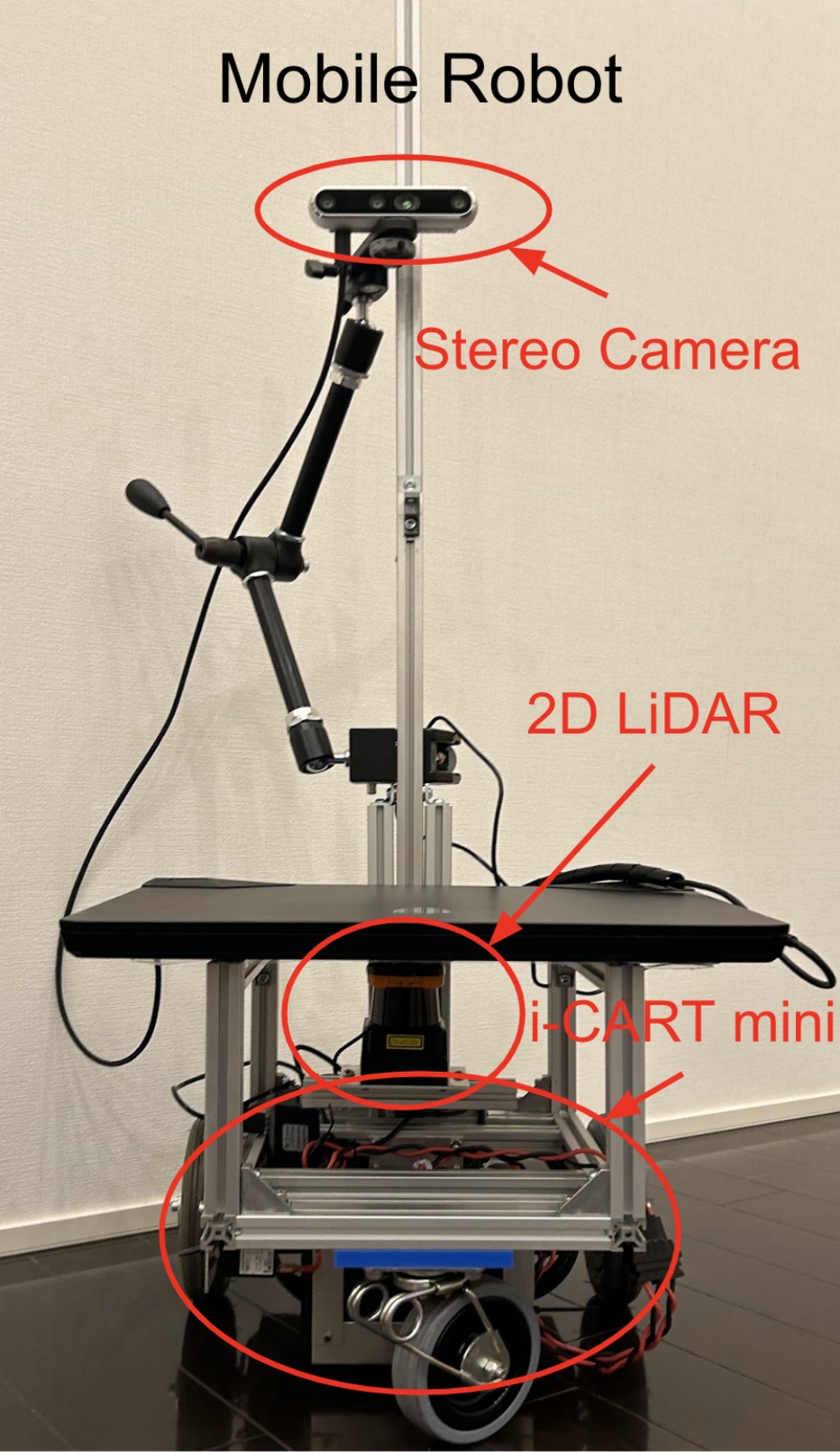}
  \caption{The robot which we use in the real world experiments. We use i-CART mini for the mobile robot. The stereo camera is attached at 0.88 meters. 2D LiDAR is attached close to the floor so that SLAM can determine where the robot can navigate.}
  \label{fig:robot}
\end{figure}

\begin{figure*}[h]
  \centering
  \includegraphics[keepaspectratio, width=1.0\textwidth]{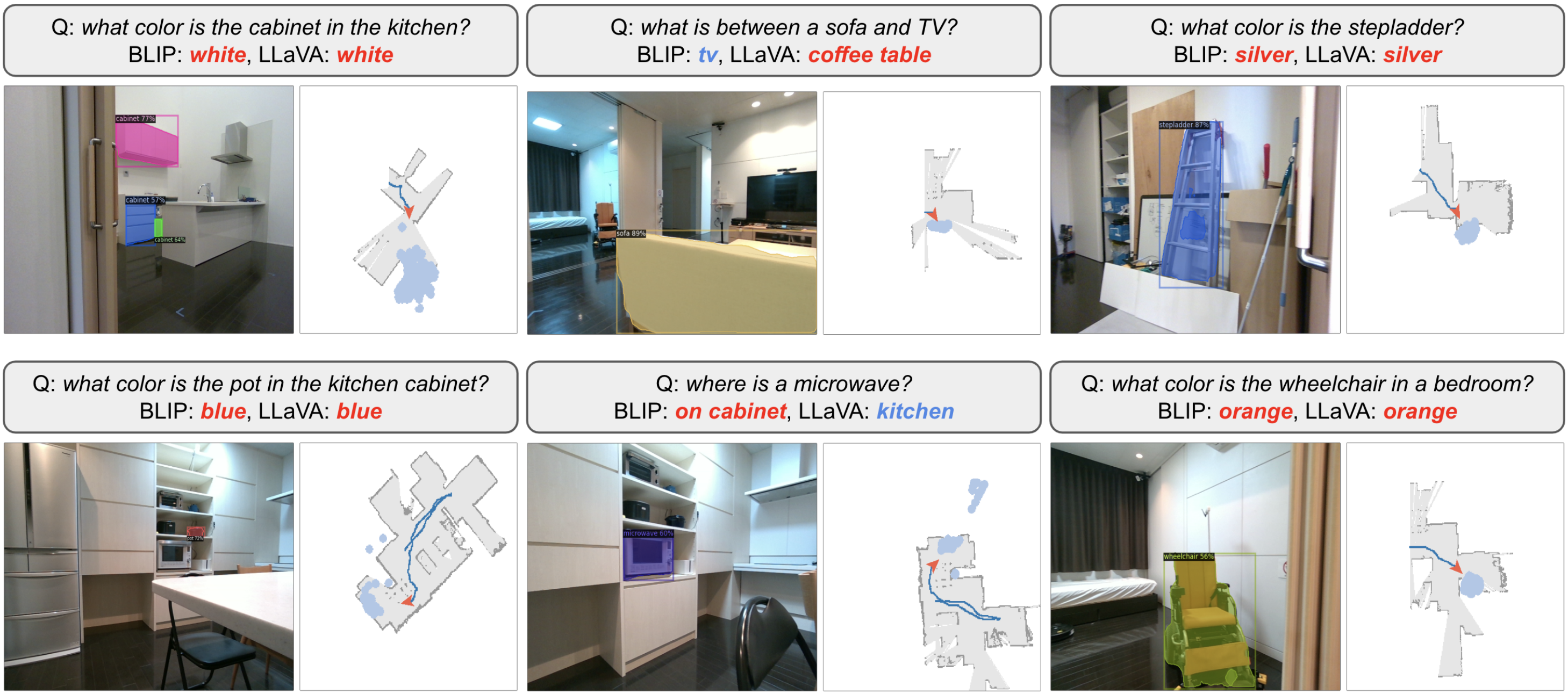}
  \caption{Qualitative examples of the success results in the real world.}
  \label{fig:Success}
\end{figure*}
\begin{figure}[h]
  \centering
  \includegraphics[keepaspectratio, width=0.49\textwidth]{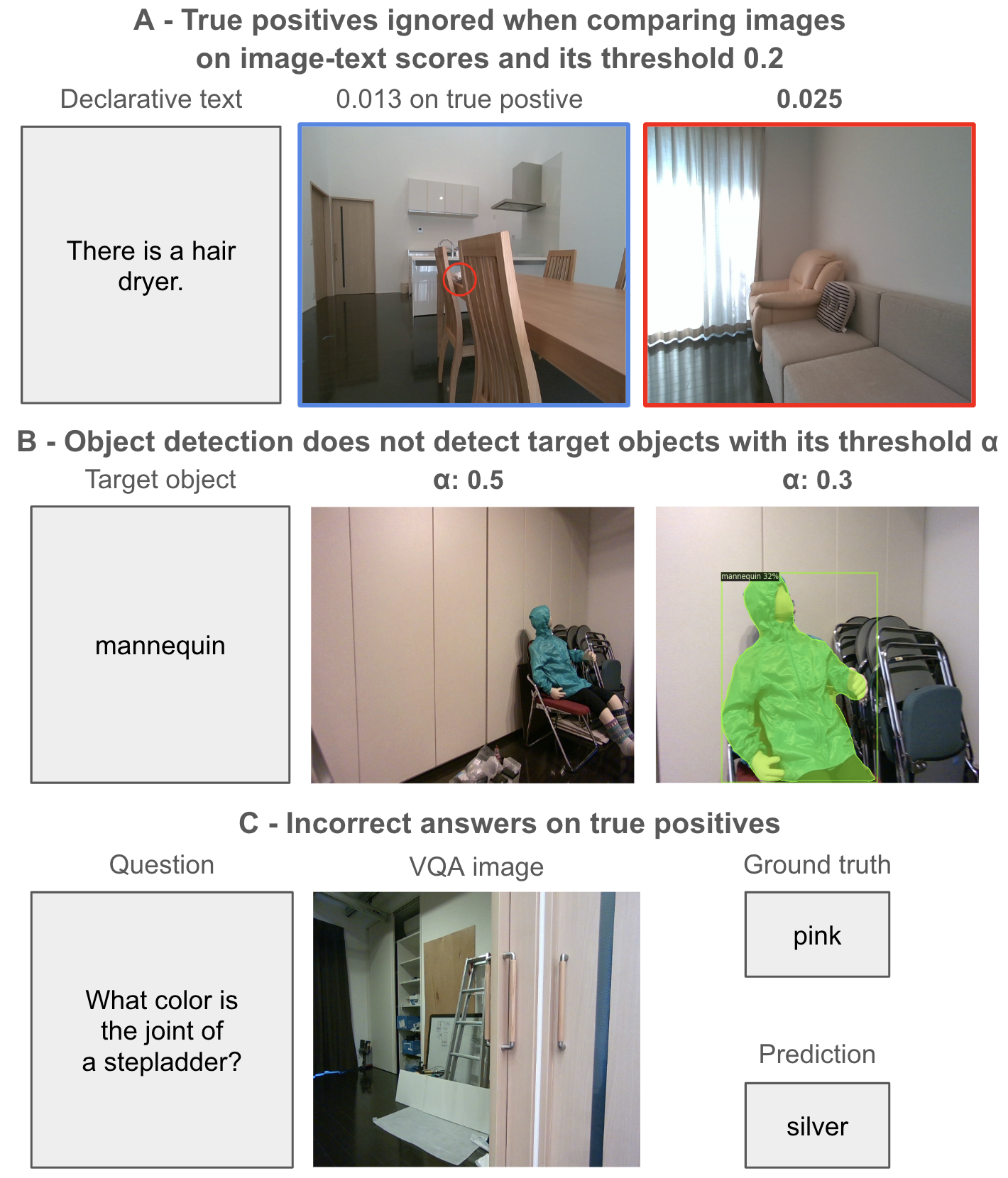}
  \caption{Qualitative examples of the failure results in the real world. (A) Matching with a threshold during exploration or comparing images post exploration can result in false negatives. (B) The semantic segmentation model cannot detect some objects with the agent’s positions and the threshold. (C) VQA model outputs incorrect answers on true positive images.}
  \label{fig:Failure}
\end{figure}

\vspace{0.2cm}
\noindent \textbf{EQA in the Real-World Environment.}
We evaluate our method in real-world environments. 
Fig.~\ref{fig:robot} shows a robot equipped with a stereo camera and 2D LiDAR. 
The robot performs 19 episodes in two houses.
We set $(\alpha, \beta)=(0.5,0.2)$ from some trials.
The robot is placed 0.5 meters (the minimum range of SLAM) away from objects and walls for initialization.
We define success in real-world settings as 
predicting the correct answer regarding a target object without any collisions.

We observe that navigation time per episode ranges from 20 seconds to 6 minutes, with a success rate of around 32 across 19 episodes shown in Table~\ref{tab:RealWorld}.
Changing image-text matching models had minimal impact on performance. 
The agent occasionally collides with objects due to our predefined rules and limitations of 2D SLAM.
When the deterministic local policy fails to generate a path to the predicted long-term goal, the agent moves randomly, increasing the likelihood of collisions.
Furthermore, 2D SLAM only maps areas that reflect a laser from the 2D LiDAR, potentially leaving some obstacles undetected.
This 2D LiDAR limitation can be resolved by using the depth of the RGB-D camera.

Figures~\ref{fig:Success} and~\ref{fig:Failure} show success and failure cases of EQA. 
Our method identifies the target object even among multiple similar objects, but failures occur in four areas: navigation, image-text matching, object detection, and VQA. 
Navigation failures arise when the agent can't find the target within the given steps. 
In image-text matching, the correct image may be discarded if its score is lower than others. 
Object detection errors, such as failing to detect mannequins or clothes, lead the agent to continue exploring. 
VQA failures occur when incorrect answers are given despite having the correct image. 
Our method also struggles with counting objects, especially when items like chairs are spread across multiple rooms.
A more advanced memory architecture is needed. 
Additionally, frontier-based exploration is not a efficient policy as the agent has more information about the target object such as colors and locations.
Better exploration can be considered our future work.

\section{CONCLUSIONS}

In this paper, we propose a map-based modular approach for zero-shot EQA, combining ObjNav and VQA modules. 
Through extensive experiments in both virtual and real-world settings, we demonstrate that our approach outperforms the VQA-only baseline, suggesting that our navigation and memory architecture contribute to the better EQA performance.
The use of a map-based navigation system allowed the agent to efficiently explore unfamiliar spaces and locate target objects, while the integration of vision-language models like BLIP2 and LLaVA ensures accurate image-text matching and robust question answering.
However, certain challenges remain, particularly in the areas of navigation precision, object detection, and handling complex VQA scenarios like counting multiple objects across different rooms. 
Future work could focus on refining the navigation module to utilize more information about the target object, as well as enhancing the VQA module’s capacity for complex reasoning tasks. 
Overall, our proposed method presents a significant step forward in advancing embodied AI systems, offering a more versatile and scalable solution for real-world EQA tasks.
\section{Acknowledgements}
This work was supported by JST PRESTO Grant Number JPMJPR22P8, and JSPS KAKENHI Grant Numbers JP21K12055, JP22K12159, JP22K17983, JP22KK0184, Japan.
This research project has benefitted from the Microsoft Accelerate Foundation Models Research (AFMR) grant program through which leading foundation models hosted by Microsoft Azure along with access to Azure credits were provided to conduct the research.

\bibliographystyle{plain}
\bibliography{root}

\end{document}